\documentclass[runningheads]{llncs}

\usepackage{graphicx}
\usepackage{graphbox}
\usepackage{amsmath,amssymb} 
\usepackage{color}
\usepackage{wrapfig}
\usepackage{booktabs}
\usepackage{siunitx}
\usepackage{picinpar}

\usepackage{tikz}
\usetikzlibrary{calc}

\usepackage{siunitx}

\renewcommand{\d}{\mathrm{d}}

\begin{document}
\title{Direct Sparse Odometry with Rolling Shutter} 

\titlerunning{Direct Sparse Odometry with Rolling Shutter}

\author{David Schubert
\and
Nikolaus Demmel
\and
Vladyslav Usenko
\and \\
J\"org St\"uckler
\and 
Daniel Cremers
}
\authorrunning{D. Schubert, N. Demmel, V. Usenko, J. St\"uckler, D. Cremers}

\institute{Technical University of Munich, Garching b. M\"unchen, Germany\\
\email{\{schubdav, demmeln, usenko, stueckle, cremers\}@in.tum.de}\\
}
\maketitle
\begin{abstract}
Neglecting the effects of rolling-shutter cameras for visual odometry (VO) severely degrades accuracy and robustness.
In this paper, we propose a novel direct monocular VO method that incorporates a rolling-shutter model.
Our approach extends direct sparse odometry which performs direct bundle adjustment of a set of recent keyframe poses and the depths of a sparse set of image points.
We estimate the velocity at each keyframe and impose a constant-velocity prior for the optimization.
In this way, we obtain a near real-time, accurate direct VO method.
Our approach achieves improved results on challenging rolling-shutter sequences over state-of-the-art global-shutter VO.

\end{abstract}
\section{Introduction}

\begin{figure}[b]
    \centering
    \includegraphics[width=0.49\linewidth]{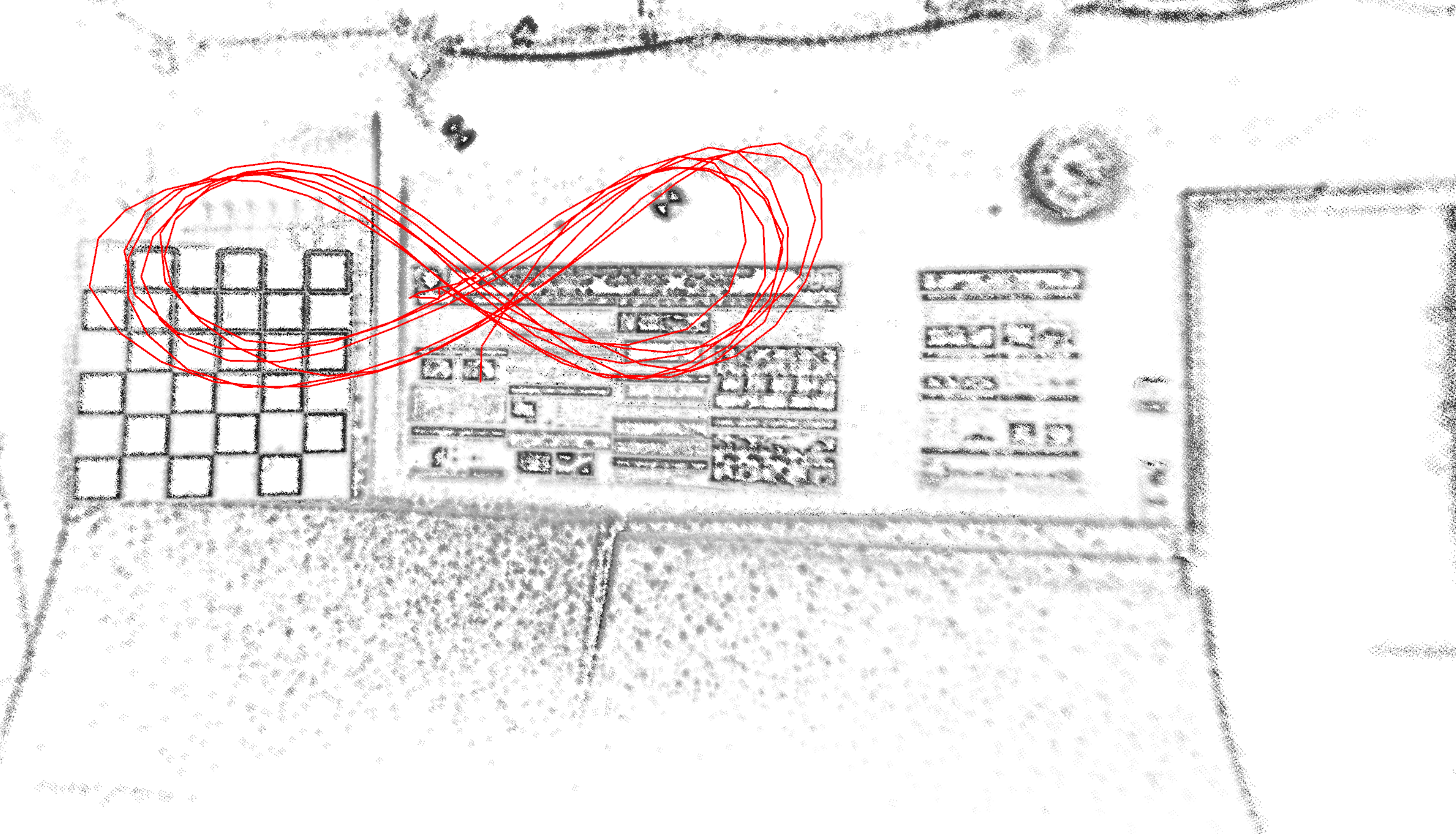}
    \includegraphics[width=0.49\linewidth]{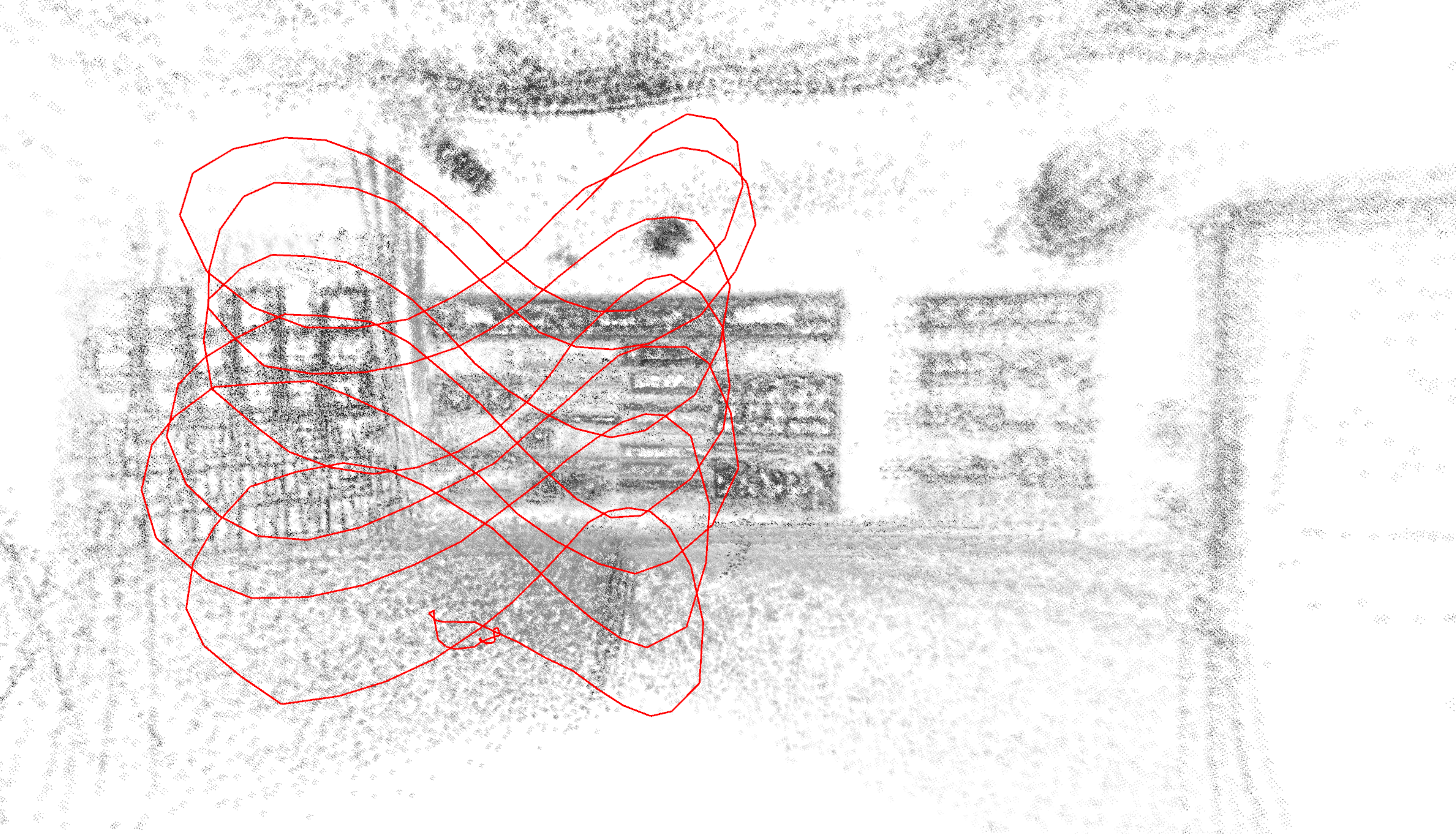}
    \caption{Qualitative result of our method (DSORS, left) versus DSO (right) on the sequence \emph{infinity-1}. The keyframe trajectory in red shows significant drift without a rolling shutter model. DSORS also produces much cleaner edges in the sparse 3D reconstruction than DSO.}
    \label{fig:qualitative}
\end{figure}

Visual odometry  for global-shutter cameras has been extensively studied in the last decades (e.g.~\cite{engel2017direct,mur2017orb}).
Global-shutter cameras capture all pixels in the image within the same time of exposure.
Most consumer grade devices such as smartphones or tablets, however, include rolling-shutter cameras which read out image rows sequentially and hence start to expose the rows at sequentially increasing capture times.
This leads to image distortions when the camera is moving. Hence, it is necessary to consider the camera pose as a function of the capture time, i.e. row index.
Simply neglecting this effect and assuming global shutter can lead to significant drift in the trajectory and 3D reconstruction estimate -- see Fig.~\ref{fig:qualitative}.

For visual odometry two major paradigms exist: direct methods (e.g.~\cite{engel2017direct}) align image intensities based on the photoconsistency assumption, while indirect methods (e.g.~\cite{mur2017orb}) align pixel coordinates of matched keypoints, i.e. they minimize keypoint reprojection error. Direct methods are particularly advantageous in weakly textured and repetitive regions. However, as demonstrated in~\cite{engel2017direct}, geometric noise as caused by neglecting rolling shutter significantly downgrades the performance of direct methods. Thus, for direct methods, it is important to model rolling shutter. 

While in indirect methods, the row and, hence, the capture time of corresponding keypoints can be assumed known through the extraction process, in direct methods, one has to impose the \emph{rolling shutter constraint} \cite{meingast2005geometric}  in order to retrieve the capture time. 
In this paper, we propose a novel direct visual odometry method for rolling-shutter cameras.
Our approach performs direct bundle adjustment of a set of recent keyframe poses and the depths of a sparse set of image points.
We extend direct sparse odometry (DSO,~\cite{engel2017direct}) by estimating the velocity at each keyframe, imposing a constant-velocity prior for the optimization and incorporating rolling-shutter into the projection model.

We evaluate our method on challenging datasets recorded with rolling-shutter cameras and compare our method to a state-of-the-art approach for global shutter cameras, demonstrating the benefits of modeling rolling shutter adequately in our method.

\section{Related Work}\label{sec:relwork}

{\bf Indirect methods:} The vast set of literature on indirect methods for visual odometry and SLAM considers global-shutter cameras~\cite{nister2004_vo,mur2017orb}. Some approaches investigate the proper treatment of rolling-shutter effects.
Hedborg et al.~\cite{hedborg2012rolling} propose rolling shutter bundle adjustment which assumes constant-velocity motion between camera poses to determine the camera pose for the row of a keypoint through linear interpolation.
Insights into degenerate cases of rolling-shutter bundle adjustment are given in~\cite{albl2016degeneracies}. Essentially, 3D reconstructions collapse to a plane when the camera frame directions are parallel in which the shutter is traversing.
Ait-Aider et al.~\cite{aitaider2009_rsstereo} recover 3D reconstruction and motion of a rigidly moving object using a snapshot of a rolling-shutter stereo camera. The method assumes linear motion and solves a non-linear system of equations resulting from the keypoint correspondences. They argue that the use of rolling-shutter cameras can be beneficial over global-shutter cameras for kinetics estimation of fast moving objects. Another line of work addresses the problem of recovering pose and motion in the case of known structure from a single rolling-shutter image~\cite{ait2006simultaneous,magerand2010generic,magerand2012global,albl2015r6p}.
Dai et al.~\cite{dai2016_rsrelpose} generalize epipolar geometry to the rolling-shutter case and propose linear and non-linear algorithms to solve for the rolling-shutter essential matrix that relates two rolling-shutter images by the relative pose and rotational and translational velocities of their cameras.
Some approaches fuse vision with inertial measurements which allows for going beyond constant-velocity assumptions for inter-frame camera motion.
Lovegrove et al.~\cite{lovegrove2013spline} approximate the continuous camera motion using B-splines.
The approach of~\cite{li2013real} considers rolling-shutter for extended Kalman filter based visual-inertial odometry. Saurer et al.~\cite{saurer2016sparse} develop a pipeline for sparse-to-dense 3D reconstruction that incorporates GPS/INS readings in a rolling-shutter-aware bundle adjustment, prior to performing rolling-shutter stereo to create a dense reconstruction.

{\bf Direct methods:} 
Direct methods have been recently shown to achieve state of-the-art performance for visual odometry and SLAM with global-shutter cameras~\cite{engel14eccv,engel2017direct}.
Since in direct methods, image correspondences are found through projective warping from one image to another, they are more susceptible to errors introduced by neglecting rolling-shutter effects than indirect methods. Estimating the time of projection in the other image requires the solution of the rolling-shutter constraint~\cite{meingast2005geometric}.
The constraint implicitly relates the reprojected image row of a pixel with its capture time, i.e. image row, in the other image. Meingast el al.~\cite{meingast2005geometric} develop approximations to the constraint for several special cases of camera motion.
Saurer et al.~\cite{saurer2013_rsstereo} present dense multi-view stereo reconstruction for rolling-shutter cameras including image distortion by wide-angle lenses, while they assume the camera motion known.
Kerl et al.~\cite{kerl2015dense} use B-splines to represent the trajectory estimate continuously for visual odometry with rolling-shutter RGB-D. While we propose a direct method for monocular cameras, similar to our approach they also incorporate the rolling-shutter constraint as a hard constraint by solving for the observation time in the target frame.
Most closely related to our method is the approach by Kim et al.~\cite{kim2016direct}. It extends LSD-SLAM~\cite{engel14eccv} to rolling-shutter cameras based on a spline trajectory representation. 
In contrast to our method they require depth initialization and do not incorporate lens distortion in their model. 
Their method explicitly incorporates residuals for the rolling-shutter constraint into the non-linear least squares problem by introducing variables for the capture time of each pixel while we directly solve for the capture time. 
Their implementation runs at approx. 120\,s per frame, while our method is faster by orders of magnitude.
While their approach separates tracking and mapping, we incorporate a rolling-shutter projection model into a windowed sparse direct bundle adjustment framework (DSO~\cite{engel2017direct}), and represent trajectories using camera poses and velocities at the keyframes.
This way, we achieve accurate but run-time efficient visual odometry for rolling-shutter cameras.

\section{Direct Sparse Odometry With Rolling Shutter Cameras}

We formulate visual odometry as direct bundle adjustment in a recent window of keyframes: we concurrently estimate the camera poses of the keyframes and reconstruct a sparse set of points from direct image alignment residuals (DSO~\cite{engel2017direct}).
The method comprises a visual odometry front-end and an optimization back-end. The front-end has been left unmodified compared to DSO. It provides initial parameters for the optimization back-end and is responsible for frame and point management. New frames are tracked with respect to the latest keyframe using direct image alignment assuming a constant camera pose across the image. 

The need for a new keyframe is determined based on a heuristic that takes optical flow, camera translation and exposure changes into account. The front-end also decides for the marginalization of keyframes and points which drop out of the optimization window: Keyframes are dropped if they do not have at least \SI{5}{\percent} of their points visible in the latest keyframe. Also, if the number of keyframes exceeds a threshold ($N=7$), a keyframe is selected for marginalization using a heuristic that keeps keyframes well-distributed in space, with more keyframes close to the latest one. When a keyframe is marginalized, first all points hosted in the keyframe are marginalized, then the keyframe variables are marginalized. Observations of other points visible in the marginalized keyframe are dropped to maintain the sparsity structure of the Hessian.

The point management aims at keeping a fixed number of active points in the optimization window. The method is sparse, i.e.\ it does not use all available information. Using more than 2000 image points hardly improves the tracking results of the global shutter method~\cite{engel2017direct}, and we also found for our method that the results do not justify the increase in runtime when using more points (see supplementary material). Candidate points are chosen in every new keyframe based on the image gradient, tracked in subsequent frames using epipolar line search and added to the set of active points for the bundle adjustment after old points are marginalized.

Our contribution lies in the optimization backend, where we introduce a model that explicitly accounts for rolling shutter. The energy contains residuals across the window of keyframes and is optimized with respect to all variables jointly using Gauss-Newton optimization.

\subsection{Model}

\begin{figure}
\centering

\begin{tikzpicture}

\newcommand{\resPh}[4]{
    \begin{scope}[xshift=#1, yshift=#2]
        \draw (-0.375,-0.225) rectangle (0.375,0.225);
        \node (resPh#3#4west) at (-0.375,0) {};
        \node (resPh#3#4east) at (0.375,0) {};
        \node at (0,0) {$E_{\mathbf{p}_{#3}#4}$};
    \end{scope}
}

\newcommand{\resVel}[3]{
    \begin{scope}[xshift=#1, yshift=#2]
        \draw (-0.375,-0.225) rectangle (0.375,0.225);
        \node (resVel#3east) at (0.375,0) {};
        \node at (0,0) {$E_{\mathbf{v}_{#3}}$};
    \end{scope}
}

\newcommand{\point}[3]{
    \begin{scope}[xshift=#1, yshift=#2, local bounding box=point#3]
        \draw (0,0) ellipse (0.6 and 0.25);
        \node at (0,0) {$d_{\mathbf{p}_#3}$};
    \end{scope}
}

\newcommand{\kframe}[3]{
    \begin{scope}[xshift=#1, yshift=#2, local bounding box=kf#3]
        \draw (0,0) ellipse (1.6 and 0.5);
        \draw (0.65,0) node{$a_#3,b_#3$} ellipse (0.45 and 0.3);
        \draw (-0.8,0) node{$\mathbf{v}_#3$} ellipse (0.3 and 0.3);
        \draw (-0.15,0) node{$\mathbf{T}_#3$} ellipse (0.3 and 0.3);
        \node (v#3) at (-1.1,0) {};
        \node (T#3north) at (-0.15,0.3) {};
        \node (T#3south) at (-0.15,-0.3) {};
    \end{scope}
}

\draw[fill=gray!15,draw=none] (-5,-0.75) rectangle (6,0.75);
\draw[fill=gray!15,draw=none] (-5,2.25) rectangle (6,3.75);
\node at (-4.5, 0) {KF3};
\node at (-4.5, 1.5) {KF2};
\node at (-4.5, 3) {KF1};

\resPh{3cm}{3.25cm}{1}{0}
\resPh{3cm}{2.75cm}{1}{2}
\resPh{3cm}{2cm}{2}{1}
\resPh{3cm}{1.5cm}{2}{3}
\resPh{3cm}{1cm}{3}{3}
\resPh{3cm}{0.25cm}{4}{2}
\resPh{3cm}{-0.25cm}{4}{4}

\point{5cm}{0cm}{4}
\point{5cm}{1.2cm}{3}
\point{5cm}{1.8cm}{2}
\point{5cm}{3cm}{1}

\kframe{0cm}{0cm}{3}
\kframe{0cm}{1.5cm}{2}
\kframe{0cm}{3cm}{1}

\resVel{-3cm}{0}{3}
\resVel{-3cm}{1.5cm}{2}
\resVel{-3cm}{3cm}{1}

\draw (kf1.east) -- (resPh10west.center);

\draw (kf1.east) -- (resPh10west.center);
\draw (kf1.east) -- (resPh12west.center);
\draw (kf1.east) -- (resPh21west.center);

\draw[shorten <= 1.1cm, dotted] ($(kf1.east)+(0,1.5cm)$) -- (resPh10west.center);
\draw[shorten <= 1.1cm, dotted] ($(kf3.east)+(0,-1.5cm)$) -- (resPh44west.center);

\draw (kf2.east) -- (resPh12west.center);
\draw (kf2.east) -- (resPh12west.center);
\draw (kf2.east) -- (resPh21west.center);
\draw (kf2.east) -- (resPh21west.center);
\draw (kf2.east) -- (resPh23west.center);
\draw (kf2.east) -- (resPh33west.center);
\draw (kf2.east) -- (resPh42west.center);

\draw (kf3.east) -- (resPh23west.center);
\draw (kf3.east) -- (resPh33west.center);
\draw (kf3.east) -- (resPh42west.center);
\draw (kf3.east) -- (resPh44west.center);

\draw (point1.west)--(resPh10east.center);
\draw (point1.west)--(resPh12east.center);

\draw (point2.west)--(resPh21east.center);
\draw (point2.west)--(resPh23east.center);

\draw (point3.west)--(resPh33east.center);

\draw (point4.west)--(resPh42east.center);
\draw (point4.west)--(resPh44east.center);

\draw (v1.center) -- (resVel1east.center);
\draw (v2.center) -- (resVel2east.center);
\draw (v3.center) -- (resVel3east.center);

\draw (T3north.center) to[out=90, in=8] (resVel3east.center);
\draw (T2south.center) -- (resVel3east.center);

\draw (T2north.center) to[out=90, in=8] (resVel2east.center);
\draw (T1south.center) -- (resVel2east.center);

\draw (T1north.center) to[out=90, in=8] (resVel1east.center);
\draw[shorten <= 1.2cm, dotted] ($(T1south.center)+(0,1.5cm)$) -- (resVel1east.center);

\end{tikzpicture}

\caption{Factor graph of the objective energy. The $i$th point observed in the $j$th keyframe contributes a photometric energy $E_{\mathbf{p}_ij}$, which depends on the variables of the host and the target keyframe plus the point's inverse depth in the host frame. In addition, energy terms $E_{\mathbf{v}_i}$ representing the velocity prior create correlations between velocities and poses. Not shown are the camera intrinsics, which influence every photometric residual.}
\label{fig:graph}
\end{figure}
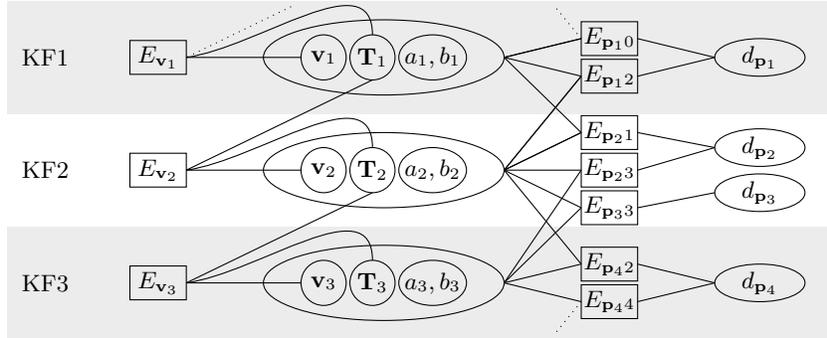

In the following, we detail our formulation of direct image alignment in the optimization backend of DSO for rolling shutter cameras. As the rows of an image are not taken at the same time, it is now necessary to find the camera pose as a function of time $t$. Camera poses $\mathbf{T}_i(t)$ are elements of the special Euclidean group $\mathrm{SE}(3)$. We choose a constant velocity model, such that the parametrization of the camera motion while frame $i$ is being taken is given by

\begin{align}
\mathbf{T}_i(t)&=\exp(\hat{\mathbf{v}}_i t)\mathbf{T}_{0,i},
\end{align}
where $\mathbf{v}_i\in\mathbb{R}^6$ is a velocity vector that includes both translational and rotational components and $\hat{\mathbf{v}}_i\in\mathfrak{se}(3)\subset\mathbb{R}^{4\times 4}$ the corresponding Lie Algebra element. We assume that the time $t$ when a pixel is read out is linearly related to the vertical $y$-coordinate of the pixel, i.e.
\begin{align}
t(x,y)=y-y_0\label{eq:rowtime}
\end{align}
which usually is well satisfied for rolling shutter cameras (though technically the shutter might also be along the horizontal $x$-coordinate, in which case the image can simply be rotated). The optimization backend estimates the reference poses~$\mathbf{T}_{0,i}\in\mathrm{SE}(3)$ of the keyframes and the velocities $\mathbf{v}_i\in\mathbb{R}^6$.

In our model, the row $y_0$ has been taken at the camera pose $\mathbf{T}_{0,i}$.
We set $y_0$ in the middle of the vertical range of the image. Of course, $y_0$ can be chosen arbitrarily in the image and we could just choose $y_0=0$, but with our choice we assume that the optimal $\mathbf{T}_{0,i}$ is in best agreement with its initialization from tracking towards the last keyframe which assumes global shutter.

If a point $\mathbf{p}$ that is hosted in image $I_i$ is observed in image $I_j$, it contributes to the energy as
\begin{align}
E_{\mathbf{p}j} &= \sum_{k\in\mathcal{N}_\mathbf{p}} w_{\mathbf{p}_k} \| r_k \|_\gamma\,,
\end{align}
with photometric residuals
\begin{align}
    r_k = (I_j[\mathbf{p}_k']-b_j) - e^{a_j-a_i}(I_i[\mathbf{p}_k]-b_i).
\end{align}
The indices in $\mathcal{N}_\mathbf{p}$ denote pixels in the neighborhood of point $\mathbf{p}$. As in \cite{engel2017direct}, we use an 8-pixel neighborhood and a gradient-based weighting $w_{\mathbf{p}_k}$ that down-weights pixels with strong gradient. For robustness, the Huber norm $\|\cdot\|_\gamma$ is used. The parameters $a_i,b_i$ describe an affine brightness transfer function $\exp(-a_i)(I_i-b_i)$ which is used to account for possibly changing exposure times or illumination. For photometrically perfect images (synthetic data), they are not required. For the real data in our experiments with constant exposure we can include a prior which keeps them at zero.

The pixel position $\mathbf{p}_k'$ in the host frame is calculated from $\mathbf{p}_k$ with a composition of inverse projection, rigid body motion and projection,
\begin{align}
\mathbf{p}_k' = \Pi_{\mathbf{c}}(\mathbf{R}\Pi_{\mathbf{c}}^{-1}(\mathbf{p}_k, d_\mathbf{p})+\mathbf{t})\,,
\end{align}
where $d_\mathbf{p}$ is the depth of point $\mathbf{p}$ in its host frame and the projection $\Pi_{\mathbf{c}}$ depends on the four internal camera parameters $f_x$, $f_y$, $c_x$, $c_y$ which are the components of the vector $\mathbf{c}$.

The rotation $\mathbf{R}$ and the translation $\mathbf{t}$ are calculated as
\begin{align}
\begin{bmatrix}
\mathbf{R} & \mathbf{t} \\ \mathbf{0} & 1
\end{bmatrix}
&= \mathbf{T}_j(t^*)\mathbf{T}_i^{-1}(t(\mathbf{p})) \\
&=\exp(\hat{\mathbf{v}}_j t^*)\mathbf{T}_{0,j}\mathbf{T}_{0,i}^{-1}\exp(-\hat{\mathbf{v}}_i t(\mathbf{p}))
\end{align}
Note that we know the time when the point has been observed in the host frame, as we know its pixel coordinates. It is not straightforward, however, to obtain the time $t^*$ of observation in the target frame. It depends on the $y$-coordinate $\mathbf{p}_y'$ of the projected point (eq.~\eqref{eq:rowtime}), but the $y$-coordinate of the projected point also depends on time through the time-dependent pose $\mathbf{T}_j(t)$. This interdependency is formulated as the \emph{rolling shutter constraint} \cite{meingast2005geometric}, where we choose $t^*$ such that it satisfies
\begin{align}\label{eq:rsc}
t^* = t(\mathbf{p}'(t^*)).
\end{align}
In short: pose time equals row time. Here, $t$ is the function defined in eq.~\eqref{eq:rowtime}. Apart from some specific cases, there is no closed-form solution for $t^*$~\cite{meingast2005geometric}. In practice, it turns out to be sufficient to iterate the update $t^* \gets t(\mathbf{p}'(t^*))$ for a few times to obtain a solution.
 
We remove lens distortion from the images through undistortion in a preprocessing step. Consequently, the mapping between pixel coordinates and time in eq.~\eqref{eq:rowtime} changes: Instead of the row in the undistorted image, we need to consider the row of the corresponding point in the original distorted image,
\begin{align}
t(x,y) = f_{\text{d}}(x,y)_y-\tilde{y}_0.
\end{align}
A point in the undistorted image with coordinates $x$ and $y$ is mapped by the distortion function $f_\text{d}$ into the original image, where the $y$-coordinate determines the time. The offset $\tilde{y}_0 = f_{\text{d}}(x_0,y_0)_y$ is chosen to be the original $y$-coordinate of the midpoint $(x_0, y_0)$ of the undistorted image.
We calculate $f_{\text{d}}(x,y)_y$ for all pixels as a preprocessing step and then interpolate. This is computationally less expensive than using the distortion function each time. Also, it facilitates the incorporation of new distortion models, as we later need the derivatives of the function.

The first component of the total energy is the summation over all photometric residuals,
\begin{align}\label{eq:phenergy}
E_\text{ph} = \sum_{i\in\mathcal{F}}\sum_{\mathbf{p}\in\mathcal{P}_i}\sum_{j\in\text{obs}(\mathbf{p})}E_{\mathbf{p}j},
\end{align}
where $\mathcal{F}$ is the set of frames, $\mathcal{P}_i$ the set of points in frame $i$ and $\text{obs}(\mathbf{p})$ the set of frames in which $\mathbf{p}$ is visible.

Optimizing $E_\text{ph}$ alone, however, is not reliable. It has been shown that rolling shutter images are prone to ambiguities \cite{albl2016degeneracies} and we also found out that softly constraining the velocities leads to a more stable system.
Thus, we add an additional energy term $E_\text{vel}$ to the total energy,
\begin{align}\label{eq:totalcost}
E = E_\text{ph} + \lambda E_\text{vel}.
\end{align}
The term
\begin{align}
E_\text{vel} = \sum_{i\in\mathcal{F}}\|\mathbf{v}_i- \mathbf{v}_{i,\text{prior}}\|^2
\end{align}
is a prior on the velocities, with
\begin{align}
\mathbf{v}_{i,\text{prior}} = {\log(\mathbf{T}_i^{-1}\mathbf{T}_{i-1})\text{\v{}}} \frac{\Delta t_\text{r}}{t_i-t_{i-1}}\,,
\end{align}
where $\log(\cdot)\text{\v{}}$ is the composition of the matrix logarithm and the inverse hat transform which extracts twist coordinates from Lie algebra elements, so that $\mathbf{v}_{i,\text{prior}}\in\mathbb{R}^6$. 
Here, we need actual times: $t_i-t_{i-1}$ is the time difference between the capture of the $i$th and the $(i-1)$th keyframe and $\Delta t_\text{r}$ is the time difference between two consecutive pixel rows due to the rolling shutter.
The prior intuitively favors that the velocity between the latest existing keyframe and the new keyframe is similar to the velocity while taking the new keyframe. As initially many keyframes are taken, such a smoothness assumption is reasonable. The resulting constraints are visualized in Fig.~\ref{fig:graph}, where a factor graph of the different energy terms is shown. After marginalizing a keyframe's variables, the prior still acts through the marginalization term.

\subsection{Optimization}

We minimize the cost in equation (\ref{eq:totalcost}) using Gauss-Newton optimization. The linearized system is
\begin{align}
\mathbf{H}\boldsymbol{\delta}=\mathbf{b}\,,
\end{align}
with
\begin{align}
\mathbf{H}=\mathbf{J}^T\mathbf{W}\mathbf{J}\,,\\
\mathbf{b}=-\mathbf{J}^T\mathbf{W}\mathbf{r}\,.
\end{align}
The matrix $\mathbf{J}$ is the Jacobian of the residual vector $\mathbf{r}$ for the variables to optimize. 
The diagonal weight matrix $\mathbf{W}$ contains the weights of the residuals.
The Hessian $\mathbf{H}$ has a large diagonal block of correlations between the depth variables, which makes efficient inversion using the Schur complement possible. Compared to \cite{engel2017direct}, which has 8 variables per frame (6 for the camera pose plus 2 for the affine brightness transfer function), we now additionally have 6 velocity components, which gives 14 variables per frame. In addition, there are 4 internal camera parameters $\mathbf{c}$ shared among all keyframes. Each point adds one inverse depth variable.

One single row in the Jacobian, belonging to one single pixel $\mathbf{p}_k$ in the neighborhood of $\mathbf{p}$, is given by
\begin{align}
    \mathbf{J}_k &= \frac{\partial r_k(\boldsymbol\delta\boxplus\boldsymbol{\zeta})}{\partial\boldsymbol\delta}\,.
\end{align}
The state vector $\boldsymbol\zeta$ contains all variables in the system, i.e.\ keyframe poses, velocities, affine brightness parameters, inverse depths and intrinsic camera parameters. The symbol $\boxplus$ denotes standard addition for all variables except for poses where it means $\boldsymbol\delta_\text{p}\boxplus\mathbf{T}=\exp(\hat{\boldsymbol\delta}_\text{p})\mathbf{T}$, with $\boldsymbol\delta_\text{p}\in\mathbb{R}^6$ and $\mathbf{T}\in\mathrm{SE}(3)$.

The Jacobian can be decomposed as
\begin{align}
\mathbf{J}_k=
\begin{bmatrix}
    \mathbf{J}_I \mathbf{J}_\text{geo}, &\mathbf{J}_\text{photo}
\end{bmatrix}\,.
\end{align}
The image gradient
\begin{align}
\mathbf{J}_I = \frac{\partial I_j}{\partial\mathbf{p}'_k}
\end{align}
as well as the photometric Jacobian
\begin{align}
\mathbf{J}_\text{photo}=\frac{\partial r_k(\boldsymbol\delta\boxplus\boldsymbol{\zeta})}{\partial\boldsymbol\delta_\text{photo}}
\end{align}
 are evaluated at pixel $\mathbf{p}'_k$, where $\boldsymbol\delta_\text{photo}$ corresponds to the photometric variables $a_i,b_i,a_j,b_j$.
The geometric Jacobian $\mathbf{J}_\text{geo}$ contains derivatives of the pixel location (not the intensity, as it is multiplied with $\mathbf{J}_I$) with respect to the geometric variables $\mathbf{T}_i,\mathbf{T}_j,\mathbf{v}_i, \mathbf{v}_j, d,\mathbf{c}$. It is approximated as the derivative of the central pixel $\mathbf{p}'$ for all pixels $\mathbf{p}'_k$ in the neighborhood. For $\mathbf{J}_\text{geo}$, we also have to take into account that the observation time $t^*$ depends on the geometric variables. Thus,
\begin{align}
\mathbf{J}_\text{geo} = \frac{\partial{\mathbf{p}'}}{\partial\delta_\text{geo}} + \frac{\partial{\mathbf{p}'}}{\partial t^*} \frac{\d{t^*}}{\d\boldsymbol\delta_\text{geo}}\,.
\end{align}
The derivative of a function $y(x)$ that is defined as the root of a function $R(x,y)$ is given as
\begin{align}
    \frac{\d y}{\d x} = - \frac{\partial R/\partial x}{\partial R/\partial y}\,.
\end{align}As $t^*$ is defined by equation (\ref{eq:rsc}), we can use this rule to calculate  $\frac{\d{t^*}}{\d\delta_\text{geo}}$. In our case,
\begin{align}
    R(\boldsymbol\zeta, t^*) &= t^* - t(\mathbf{p}'(\boldsymbol\zeta,t^*))\\
    &=t^*-(f_\text{d}(\mathbf{p}'(\boldsymbol\zeta,t^*))_y-\tilde{y}_0)\,,
\end{align}
so that
\begin{align}
\frac{\d{t^*}}{\d\boldsymbol\delta_\text{geo}} = \frac{\frac{\partial f_{\text{d},y}}{\partial{\mathbf{p}'}}\frac{\partial\mathbf{p}'}{\partial\boldsymbol\delta_\text{geo}}}{1-\frac{\partial f_{\text{d},y}}{\partial{\mathbf{p}'}}\frac{\partial\mathbf{p}'}{\partial t^*}}\,.
\end{align}

The Jacobians $\mathbf{J}_\text{geo}$ and $\mathbf{J}_\text{photo}$ are approximated using first-estimates Jacobians \cite{huang2009first}. This means that the evaluation point of these Jacobians does not change once a variable is part of the marginalization term, while the image gradient $\mathbf{J}_I$ and the residual $r_k$ are always evaluated at the current state. 
The authors of \cite{engel2017direct} argue that $\mathbf{J}_\text{geo}$ and $\mathbf{J}_\text{photo}$ are smooth so that we can afford evaluating them at a slightly different location for the benefit of keeping a consistent system in the presence of non-linear null-spaces such as absolute pose and scale. As the translational component of the velocity is also affected by scale ambiguity, we decide to include velocities in the first-estimate Jacobian approximation.

When variables are marginalized, we follow the procedure of~\cite{engel2017direct}: We start from a quadratic approximation of the energy that contains residuals which depend on variables to be marginalized.
Variables are marginalized using the Schur complement, which results in an energy that only depends on variables which are still active in the optimization window. 
This energy acts like a prior and can be added to the energy of active residuals. 
We also include our velocity priors in the marginalization term if they depend on a variable that is marginalized, so that velocities of active keyframes are still constrained by marginalized keyframes which are temporally and spatially close.
We refer the reader to~\cite{engel2017direct} for further details on the marginalization process.

\section{Experimental Evaluation On Real And Synthetic Data}

\begin{figure}
    \centering
    \includegraphics[width=0.49\linewidth]{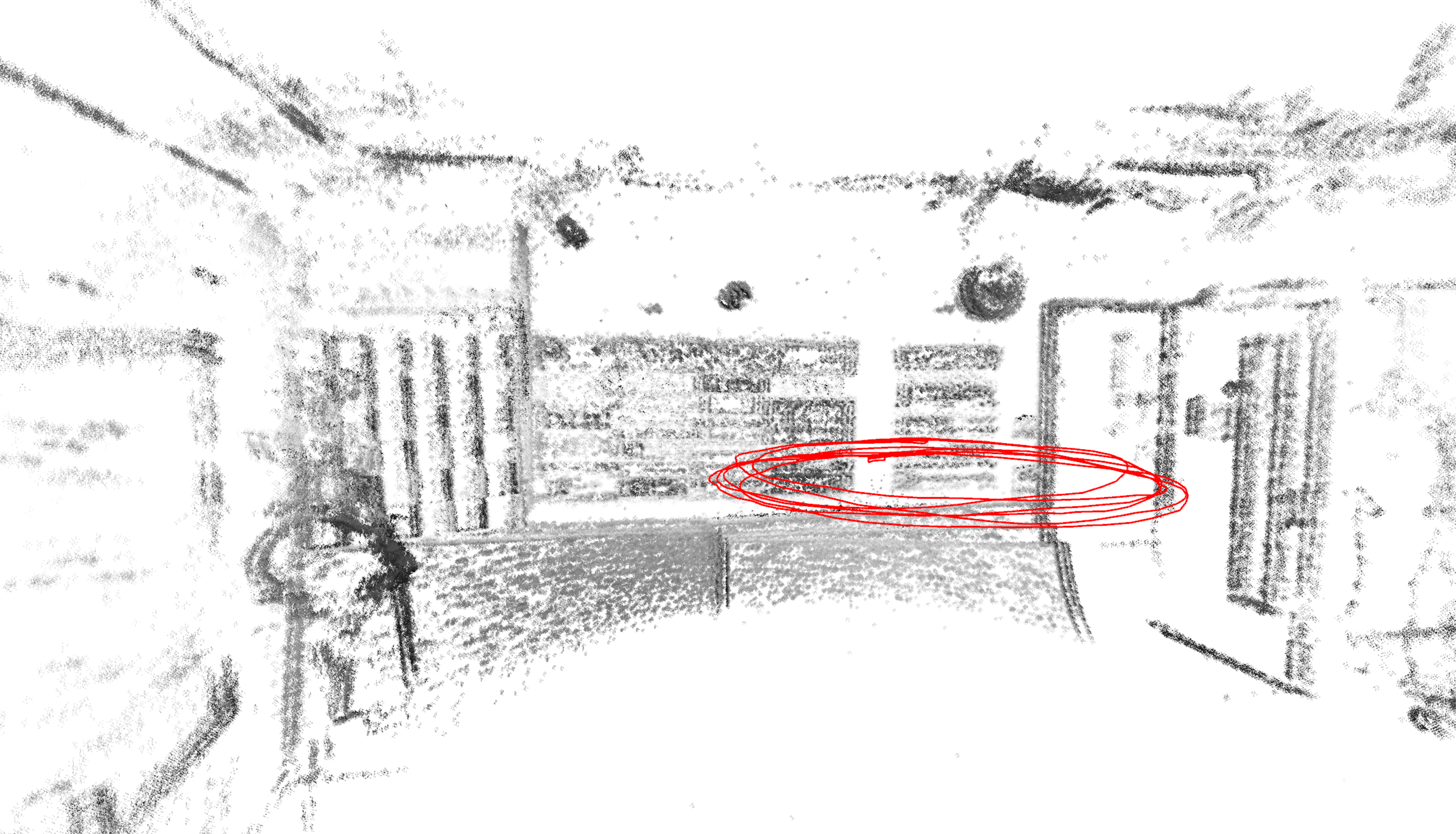}
    \includegraphics[width=0.49\linewidth]{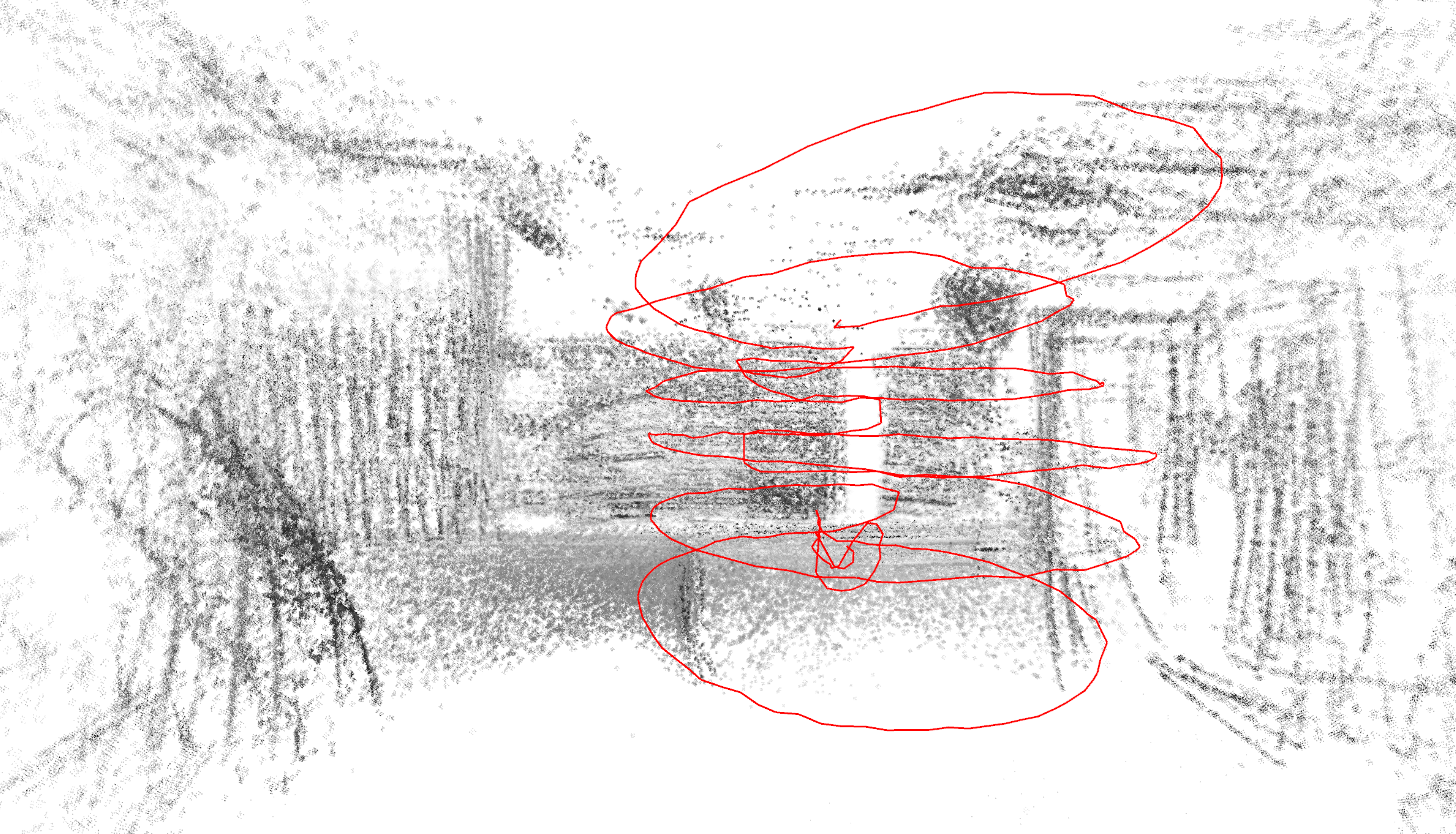}
    \caption{Qualitative result of DSORS (left) versus DSO (right) on the sequence \emph{alt-circle-1}. Even after many circles, the sparse 3D reconstruction of DSORS shows little drift, whereas DSO shows the same edges at clearly distinct locations.}
    \label{fig:qualitative14}
\end{figure}

\subsection{Datasets}

Along with the rolling shutter RGB-D SLAM method in \cite{kerl2015dense}, {\bf synthetic sequences} were published. They are re-renderings of the ICL-NUIM dataset \cite{handa2014benchmark}, containing 4 different trajectories in a living room, named \emph{kt1}, \emph{kt2}, \emph{kt3}, and \emph{kt4}. As the data is photometrically perfect, we do not estimate affine brightness parameters on these sequences.

We also show results for the sequence \emph{freiburg1\_desk}, an {\bf office sequence} from the TUM RGB-D benchmark \cite{sturm2012benchmark}. It has already been used by \cite{kim2016direct}, but a quantitative comparison is not possible, as they only show a trajectory plot. The row time difference $\Delta t_\text{r}$ is not available. We were successful with our first guess of $\Delta t_\text{r}=\SI{0.06}{ms}$, a value that leaves only a small time gap between the last row and the first row of the next image.

\begin{wrapfigure}{l}{0.25\linewidth}

    \centering
    \includegraphics[width=\linewidth]{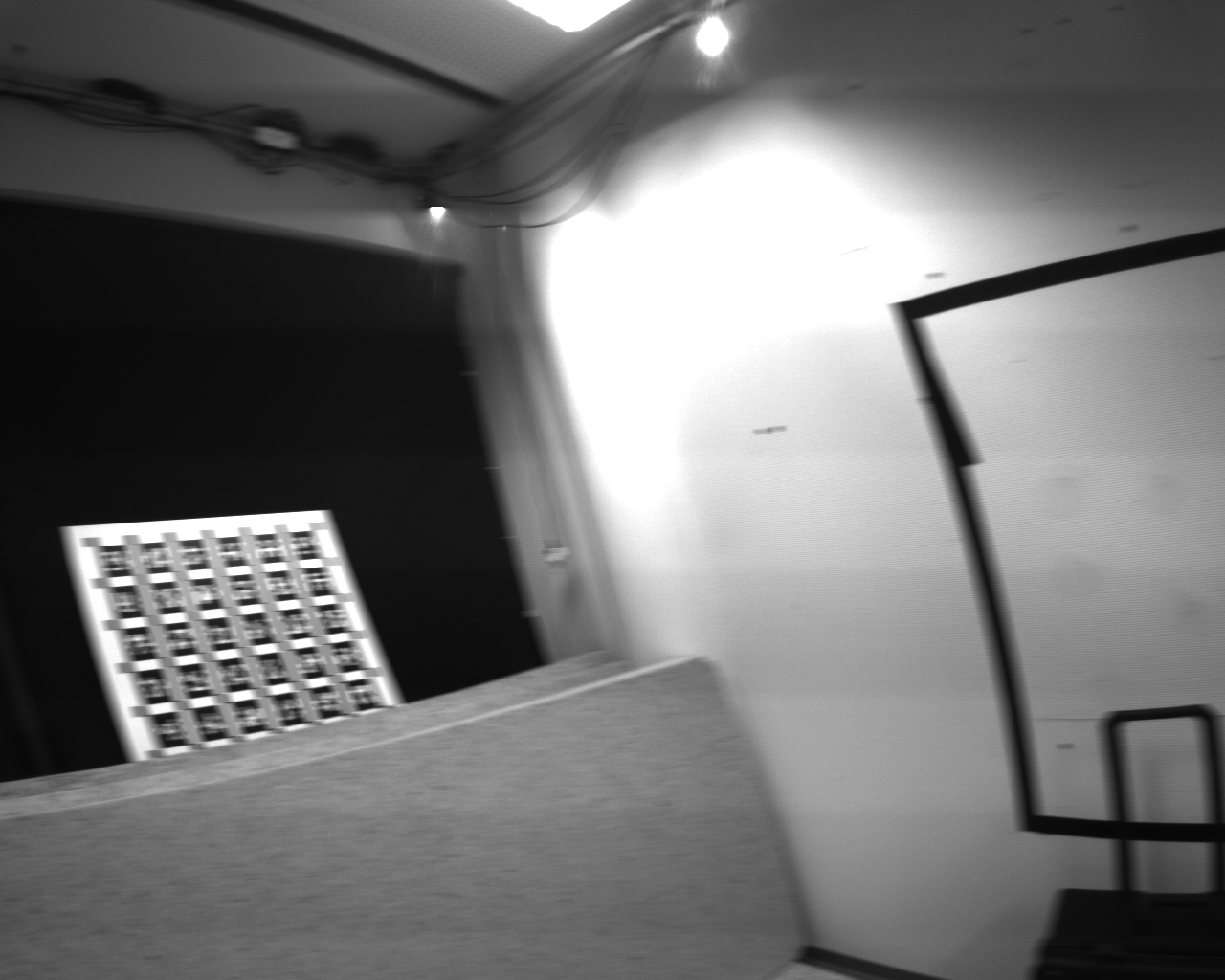}
    \caption{Our faster sequences show significant distortion and blur due to motion (from \emph{alt-circle-2}).}
    \label{fig:stripes}
\end{wrapfigure}
Due to the lack of rolling shutter datasets for monocular visual odometry, we recorded six {\bf own sequences} with ground truth. Sequences were captured at \num{1280}x\num{1024} resolution with a handheld uEye UI-3241LE-M-GL camera by IDS and a Lensagon BM4018S118 lens by Lensation. The camera was operated at \SI{20}{\hertz} and provides an approximate value of $\Delta t_\text{r}=\SI{0.033}{ms}$. Ground truth was recorded with a Flex~13 motion capture system by OptiTrack. It uses IR-reflective markers and 16 cameras distributed around the room. The ground truth poses are hand-eye and time shift calibrated, meaning that they provide poses of the camera frame to an external world system and timestamps of the ground truth poses are given in the same time system as timestamps for the camera frames. We fixed the exposure, which means our algorithm uses a prior that prefers small affine brightness parameters. We also use lens vignetting correction on our own sequences. The sequences can be divided into three categories:

\begin{itemize}
\item {\bf infinity:} The camera motion repeatedly draws an infinity symbol in front of a wall.
\item {\bf circle:} The camera moves on a circle that lies in the horizontal plane while the camera is pointing outwards
\item {\bf alt-circle:} Same as \emph{circle}, but with alternating directions.
\end{itemize}
Each of the categories has been captured twice (e.g.\ \emph{infinity-1} and \emph{infinity-2}), with the second one always being faster than the first one, but none of them is really slow in order to have sufficient rolling shutter effect.

Our camera includes two identical cameras in a stereo setup.
For comparison with methods processing global shutter images, we simultaneously recorded sequences using the second camera set to global shutter mode. 

Our dataset contains significant blur and rolling-shutter distortion due to fast motion as can be seen in Fig.~\ref{fig:stripes}.
Due to flickering of the illumination in our recording room, the rolling-shutter images contain alternating stripes of brighter and darker illumination which are also visible in Fig.~\ref{fig:stripes}.
While this is not consistent with the illumination model in DSO and DSORS, none of the two has an obvious advantage. Note that the global shutter images do not exhibit this effect. 

\subsection{Evaluation Method}

For each sequence, we compare the performance of DSORS versus DSO. DSO originally allows a maximum of 6 iterations for each Gauss-Newton optimization. We found slight improvements for our method if we increase them to 10. To make the comparison fair, we also allow a maximum of 10 iterations for DSO, though still both methods can break early when convergence is reached. The number of active points is set to \num{2000}, and there are maximally 6 old plus one new keyframe in the optimization window, which are the standard settings for DSO. Compared to DSO, we only introduced one model parameter, the weight of the velocity prior $\lambda$. The same value is used for all sequences.

We use the \emph{absolute trajectory error} (ATE) to evaluate our results quantitatively. Given ground truth keyframe positions $\hat{\mathbf{p}}_i\in\mathbb{R}^3$ and corresponding tracking results $\mathbf{p}_i\in\mathbb{R}^3$, it is defined as
\begin{align}
    e_\text{ate} = \min_{\mathbf{T}\in \mathrm{Sim}(3)}  \sqrt{\frac{1}{n}\sum_{i=1}^{n} \left\|\mathbf{T}(\mathbf{p}_i)-\hat{\mathbf{p}}_i\right\|^2}\,.
    \label{eqn:ate}
\end{align}
It is necessary to align with a 7D similarity transform\footnote{In equation (\ref{eqn:ate}) $\mathbf{T}$ is used as an operator on 3D points $\mathbf{T}: \mathbb{R}^3 \to \mathbb{R}^3, \mathbf{p} \mapsto \mathbf{T}(\mathbf{p})$.} $\mathbf{T}\in \mathrm{Sim}(3)$, since scale is not observable for monocular methods. We run the methods \num{20} times on each sequence. To randomize the results, different random number seeds are used for the point selection. We show two types of visualization for the quantitative results: the color plots show $e_\text{ate}$ for each run and for each sequence individually. The cumulative error histograms contain all sequences of each dataset. Here, the function value at position $e$ gives the number of sequences with $e_\text{ate}\leq e$.

\subsection{Results}

\begin{figure}
    \centering
    \includegraphics[width=0.49\linewidth,align=t]{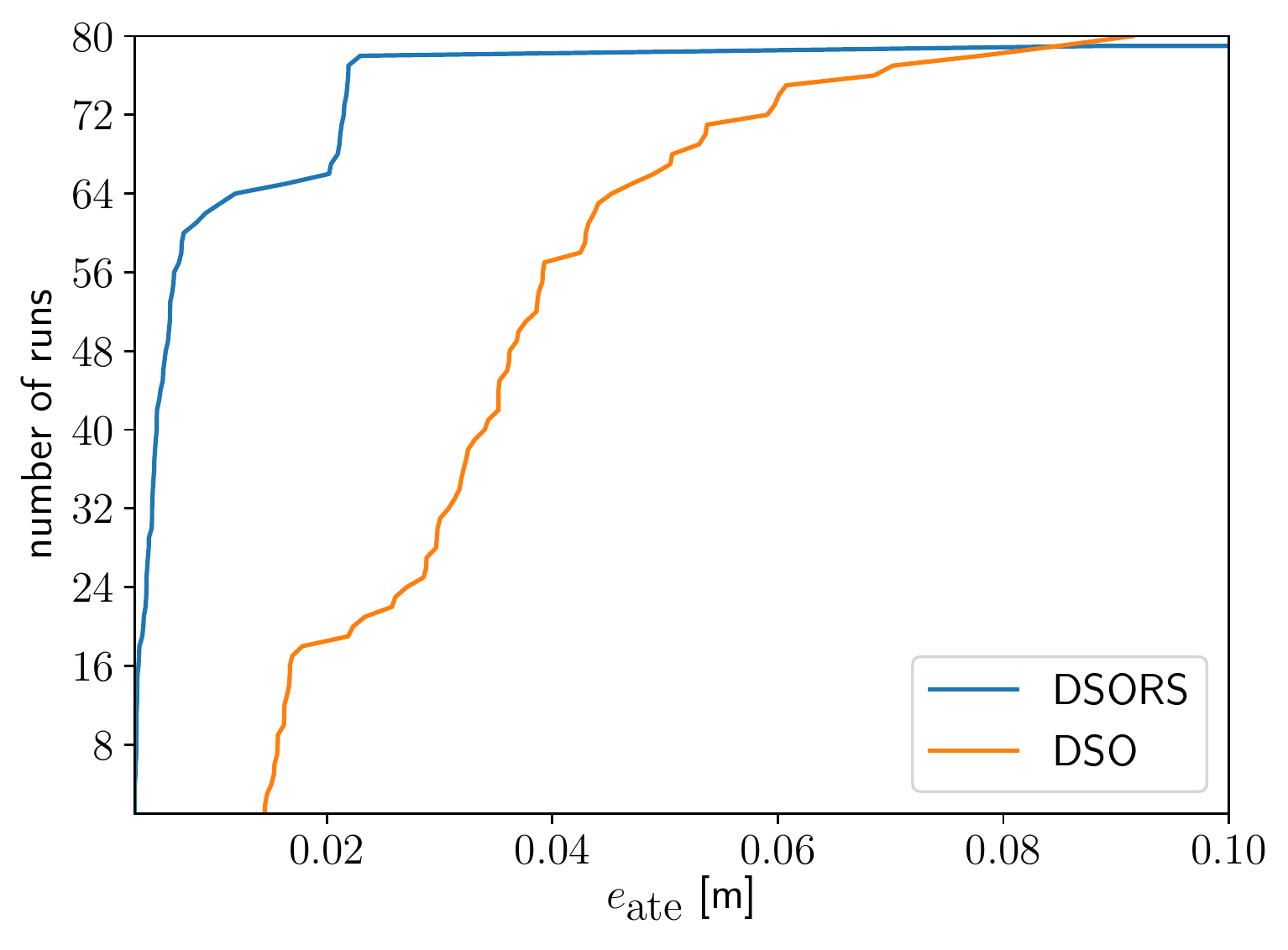}
    \includegraphics[width=0.49\linewidth,align=t]{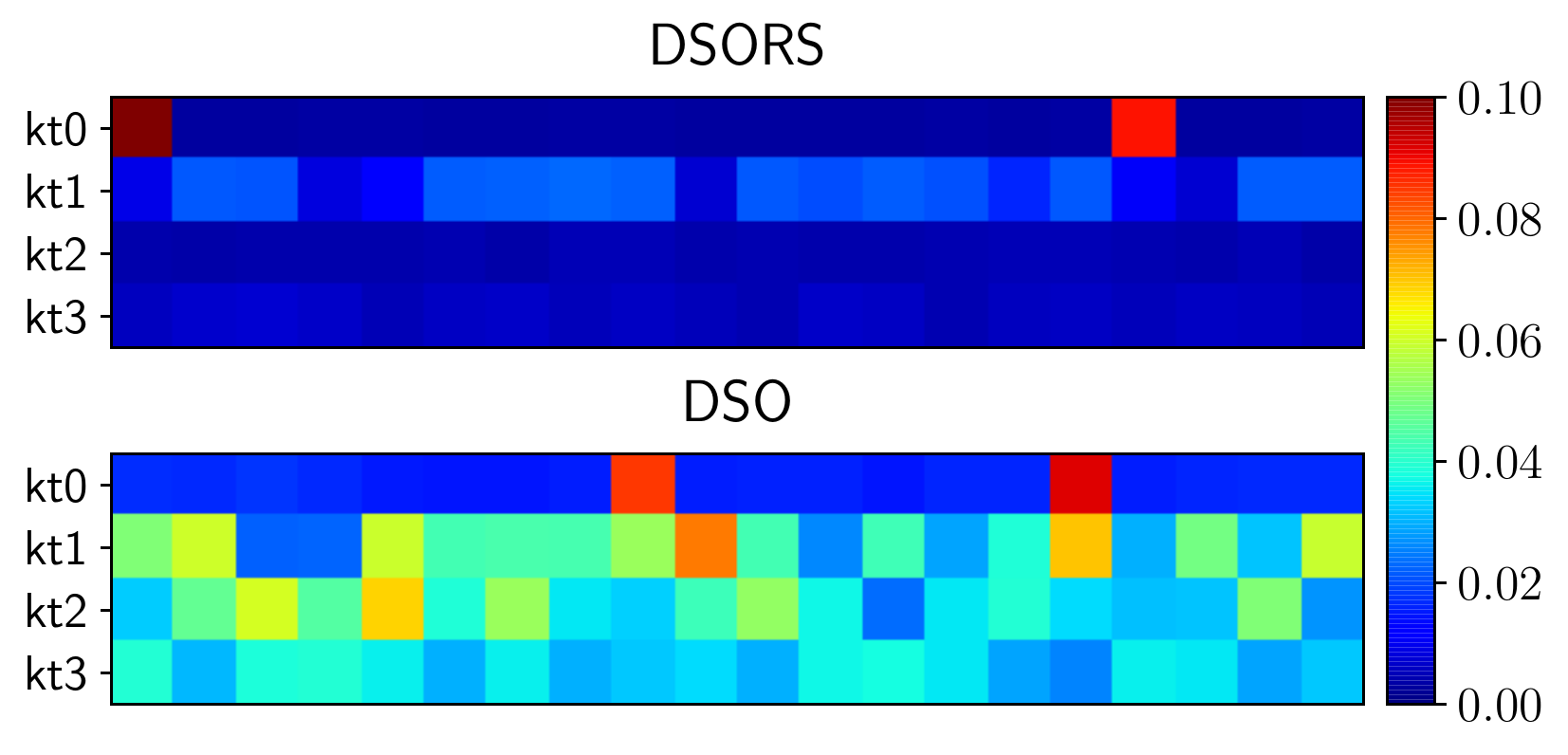}
    \caption{On the left, the cumulative error histogram for 20 runs on each synthetic sequence clearly shows that DSORS produces more accurate results than DSO. The plot on the left indicates the error for each individual run and shows that DSORS is also superior on each sequence.}
    \label{fig:resSyn}
\end{figure}
On the {\bf synthetic sequences}, DSORS clearly outperforms DSO, as can be seen in Fig.~\ref{fig:resSyn}. Not only is the overall performance in the cumulative error histogram visibly more accurate, but also on each sequence as can be seen in the color plot. Only on the sequence \emph{kt0} it is not entirely stable, but here DSO also has outliers. The RGB-D method in \cite{kerl2015dense} reports ATEs of (0.0186, 0.0054, 0.0079, 0.0210) (after $\mathrm{SE}(3)$ alignment) for the four trajectories, while our median ATEs over 20 runs are (0.0037, 0.0197, 0.0045, 0.0062) (after $\mathrm{Sim}(3)$ alignment).

\begin{figure}[!b]
    \centering
    \includegraphics[width=0.49\linewidth,align=c]{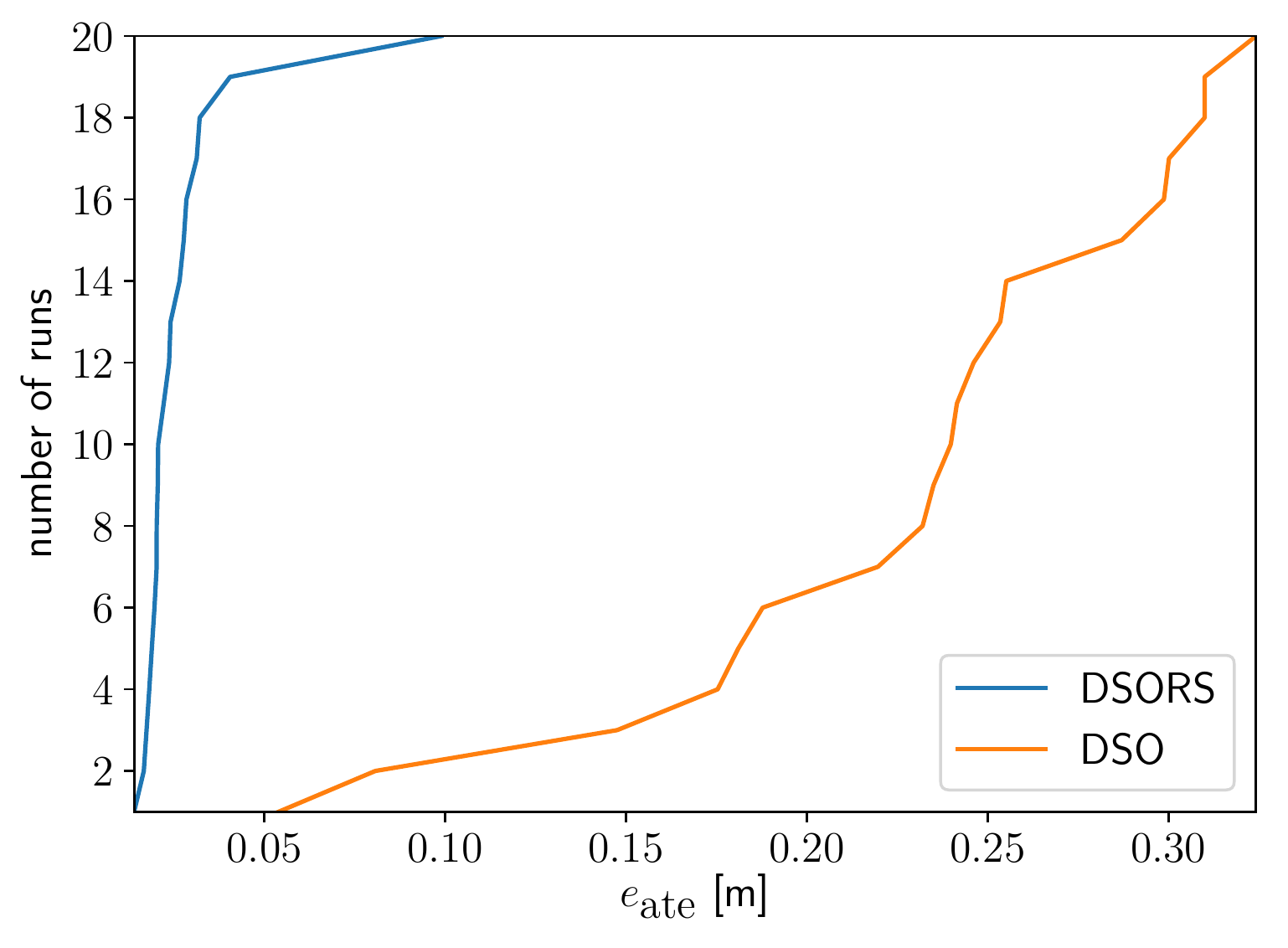}
    \includegraphics[width=0.49\linewidth,align=c]{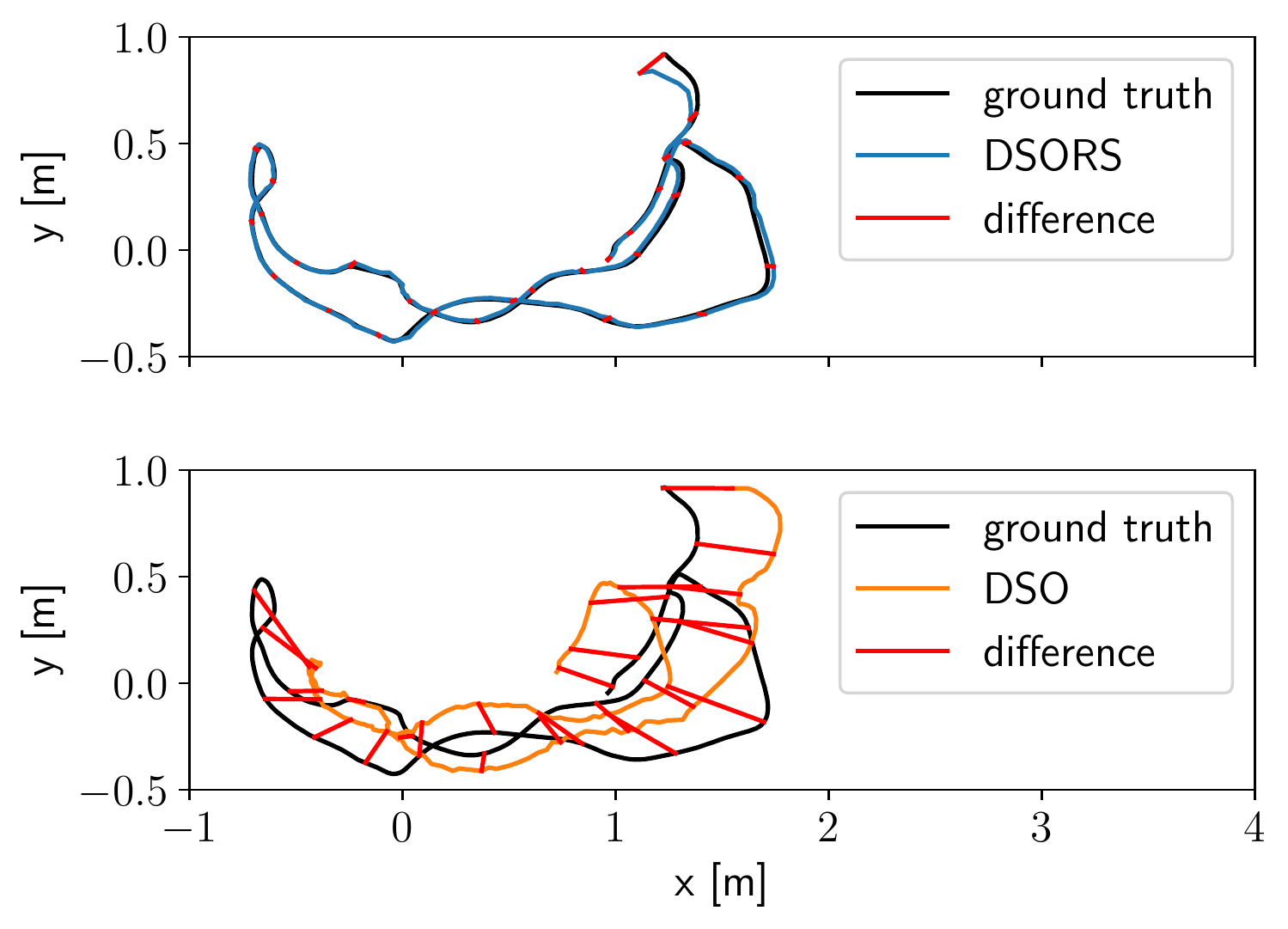}
    \caption{On the left, the cumulative error histogram over 20 iterations on the \emph{freiburg1\_desk} sequence from the TUM-RGBD benchmark shows that DSORS can repeatably track the sequence accurately while DSO has large drift. 
    On the right, the top down view shows the ground truth and estimated trajectories after $\mathrm{Sim}(3)$ alignment.
    For each method we select the iteration with median $e_\text{ate}$ (DSORS: $\SI{0.022}{m}$, DSO: $\SI{0.241}{m}$). 
    The red lines indicate corresponding points for every 10\textsuperscript{th} keyframe.}
    \label{fig:resOff}
\end{figure}

For the {\bf office sequence}, the difference between DSORS and DSO in the cumulative error histogram in Fig.~\ref{fig:resOff} is even more obvious. The performance of DSORS is much more stable and accurate. On the right side of the figure, typical tracked trajectories are plotted with ground truth after $\mathrm{Sim}(3)$ alignment. The red lines between corresponding points make visible how large the error is for DSO, compared to much smalles errors for DSORS.

The results on our {\bf own sequences} also demonstrate that DSORS is superior to DSO when dealing with rolling shutter data. Qualitatively, this is clearly visible in Figs.~\ref{fig:qualitative} and \ref{fig:qualitative14}. The sparse 3D reconstructions look much cleaner for DSORS, while DSO produces inconsistent edges when revisiting a section of the room. Even more striking is the large systematic drift of the camera trajectories.

\begin{figure}[t]
    \centering
    \includegraphics[width=0.49\linewidth,align=t]{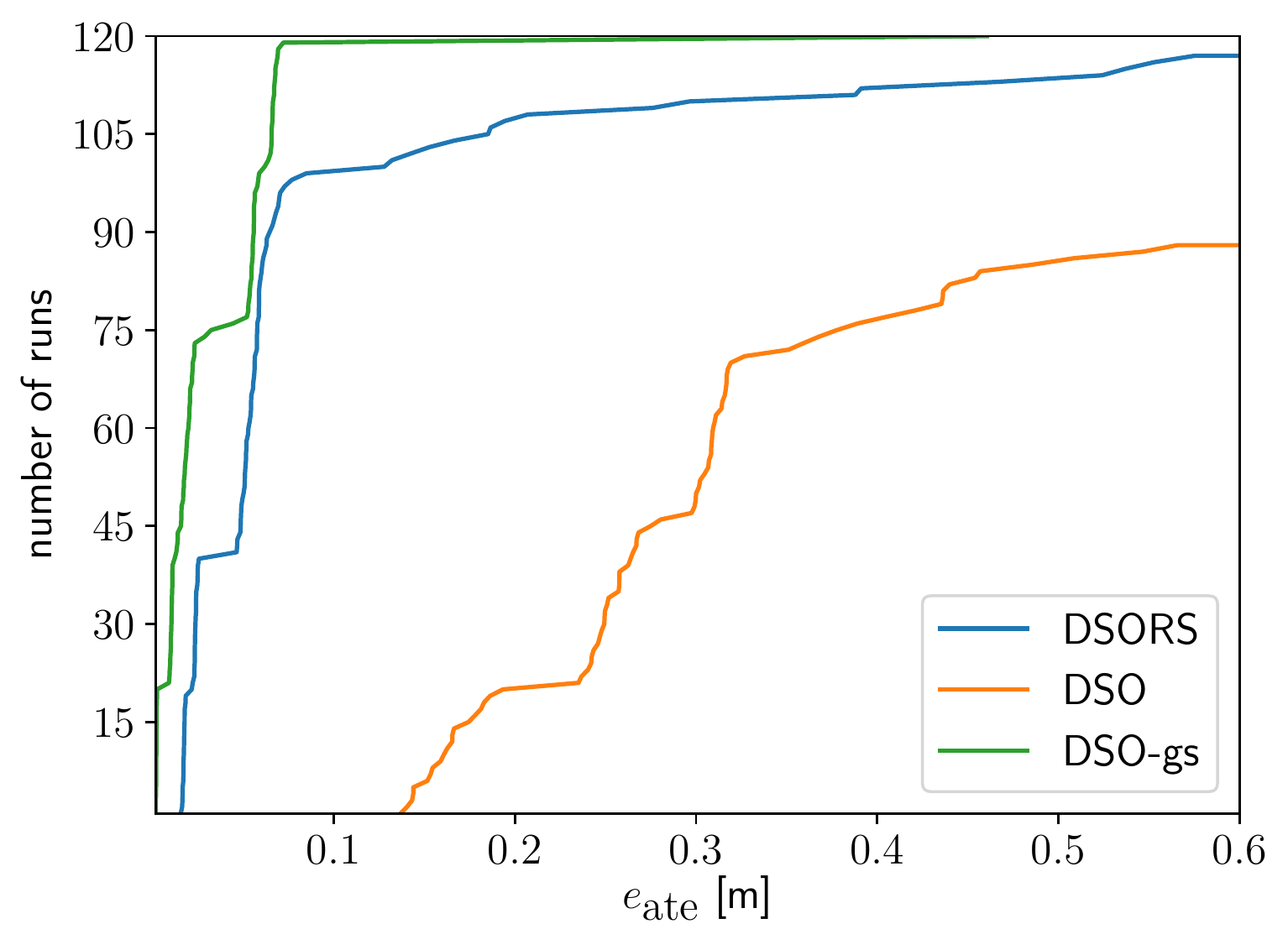}
    \includegraphics[width=0.49\linewidth,align=t]{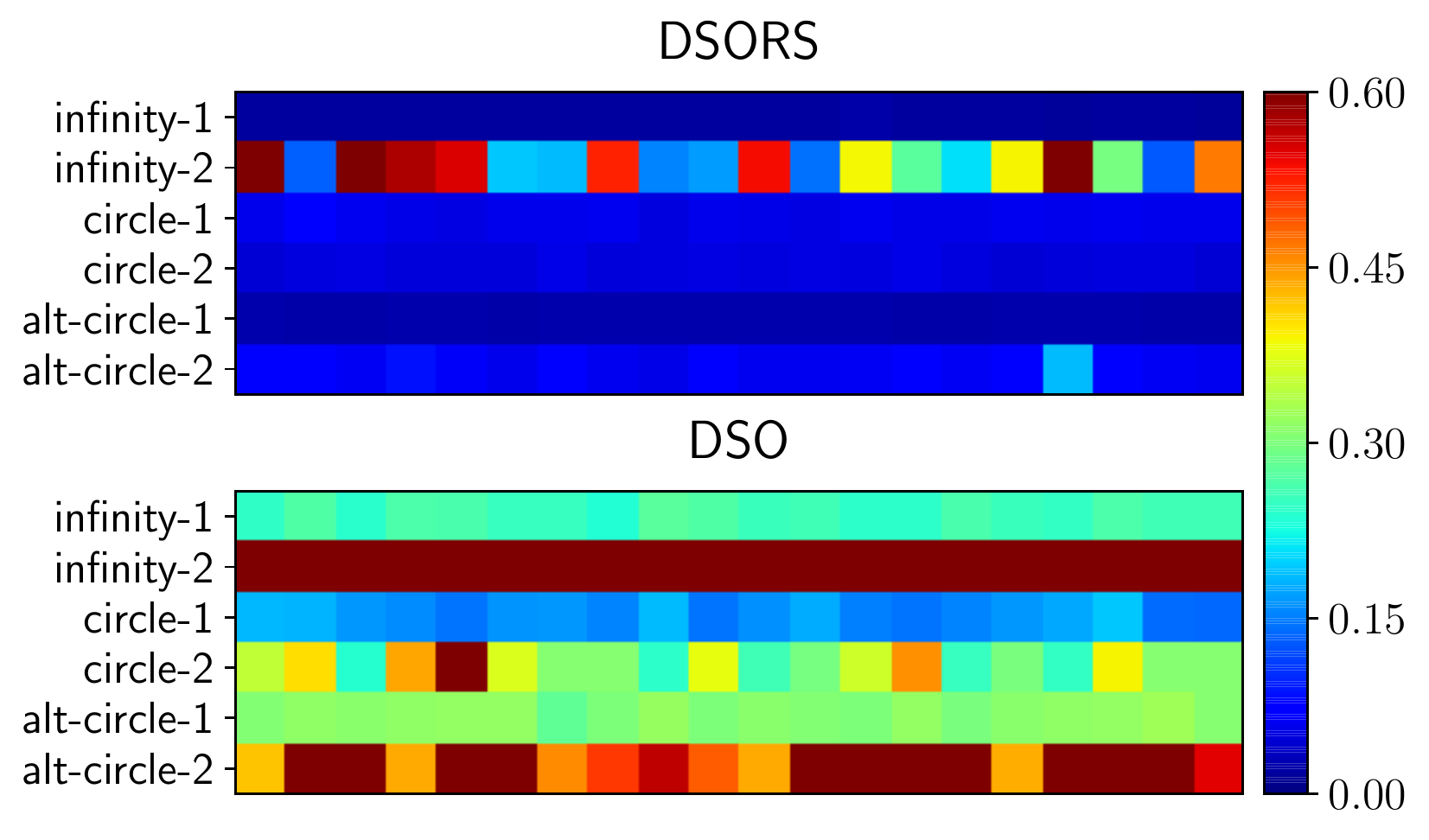}
    \caption{Cumulative error histogram and color plot for the absolute trajectory error $e_\text{ate}$ on our new sequences. In the cumulative histogram, the green line also gives a comparison to DSO running on global shutter data, which has been captured in parallel with a stereo setup. For trajectory plots as in Fig.~\ref{fig:resOff}, see supplementary material.}
    \label{fig:resOwn}
\end{figure}

In Fig.~\ref{fig:resOwn}, the quantitative difference also becomes apparent. DSORS outperforms DSO both in terms of accuracy and stability. Only the sequence \emph{infinity-2} remains a challenge, but DSORS in approximately half of the cases produces reasonable results, whereas DSO always fails in our runs.

We also show results of DSO operating on global shutter data. The sequences are very comparable to the rolling shutter sequences, as they use the same hardware, and cameras are triggered at the same time in a stereo setup. 
Running DSO on global shutter images is still better than running DSORS on rolling shutter images. 
The difference in stability visible in the cumulative histogram mainly comes from the challenging sequence \emph{infinity-2}. On the remaining sequences, the advantage of using global shutter images is not as dominant.

\subsection{Runtime}

\begin{table}[b]
\setlength{\tabcolsep}{6pt}
    \caption{Figures about trajectories and runtime. The number of keyframes that were created is denoted $n_\text{KF}$, the realtime factor is given by $r$.}
    \label{tab:statistics}
    \centering
    \begin{tabular}{llllll}
        \toprule
        Sequence & Length / Duration &  $n_\text{KF}^\text{DSORS}$ & $r^\text{DSORS}$ & $n_\text{KF}^\text{DSO}$ & $r^\text{DSO}$\\
        \midrule
        infinity-1   &  \SI{20.9}{m} / \SI{33.3}{s}    & \num{249}  & \num{0.174}&\num{240}&  \num{0.433} \\
        infinity-2       &  \SI{36.3}{m} / \SI{29.9}{s} & \num{324}  & \num{0.128}&\num{277}&  \num{0.458} \\
        circle-1       &  \SI{44.8}{m} / \SI{58.9}{s}    & \num{547}  & \num{0.149}&\num{543}&  \num{0.378} \\
        circle-2       &  \SI{30.6}{m} / \SI{29.5}{s}    & \num{424}  & \num{0.099}&\num{431}&  \num{0.248}  \\
        alt-circle-1       &   \SI{24.6}{m} / \SI{41.9}{s}   & \num{392}  & \num{0.153}&\num{399}&  \num{0.362}  \\
        alt-circle-2       &   \SI{31.6}{m} / \SI{27.1}{s}   & \num{353}  & \num{0.109}&\num{356}&  \num{0.288}  \\
        \bottomrule
    \end{tabular}
\end{table}

The computationally most expensive part of our method is the creation of a new keyframe, i.e. calculating derivatives, accumulating the Hessian and solving the linear system. Our derivatives are more involved compared to DSO, and the number of variables is larger. As keyframes are not selected with equal time spacing, but based on motion, it is not possible to specify how many frames per second we can process in general. With a fast moving camera, more new keyframes are required per time, which affects the runtime. In Table~\ref{tab:statistics}, some figures about the trajectories and the performance of DSORS and DSO run on an Intel Core i5-2500 CPU are given. The realtime factor $r$ is calculated as the real duration of the sequence divided by the processing time of the algorithm. By comparing each slower sequence (\emph{\dots-1}) to its faster variant (\emph{\dots-2}), one can confirm that the realtime factor depends on the speed of the camera motion. Only $r^\text{DSO}$ for the sequence \emph{infinity-2} is an exception, but this sequence is very unstable for DSO, thus many outlier points are dropped during the optimization, which speeds up the execution. Also, the number of keyframes is rather related to the total length of the trajectory than to the duration.

The results also prove that DSORS is slower than DSO, by a factor roughly around \num{2.5} (except for \emph{infinity-2}). It might seem surprising that not even DSO is real-time here, but this is due to the fact that all results were created in a linearized mode, where the coarse tracking waits for the keyframe creation to finish. 
Given that DSO is generally a real-time capable method, further optimization of our method or using a faster processor might produce real-time results in the future. In fact, by reducing the number of active points to \num{800} and enforcing real-time execution (with the coarse tracking continuing while a new keyframe is created), it was possible to obtain results for the slower sequences that were close to our non-real-time results, but not yet for the faster sequences.

\section{Conclusions}

In this paper, we have integrated a rolling shutter model into direct sparse visual odometry. 
By extending keyframe poses with a velocity estimate and imposing a constant-velocity prior in the optimization, we obtain a near real-time but accurate direct visual odometry method.

Our experiments on sequences from rolling shutter cameras have demonstrated that the model is well-suited and can drastically improve accuracy and stability over methods that neglect rolling shutter. 
Our method makes accurate direct visual odometry available to rolling shutter cameras which are often present in consumer grade devices such as smartphones or tablets and in the automotive domain.

For direct rolling shutter approaches, real-time capability is a challenge. With our formulation, we are already much closer to real-time processing than the alternative approach in \cite{kim2016direct}. In future work, we will investigate further speed-ups of our implementation. The integration with inertial sensing could further increase the accuracy and stability of the visual odometry.

\section*{Acknowledgment}
This work was partially supported through the grant ``For3D" by the Bavarian Research Foundation and through the grant CR~250/9-2 ``Mapping on Demand" by the German Research Foundation.

\bibliographystyle{splncs04}
\bibliography{new}
\end{document}